\documentclass[11pt]{article}
	
	\newcommand{\blind}{0}
	
    \makeatletter
    \renewcommand\section{\@startsection {section}{1}{\z@}%
                                       {-3.5ex \@plus -1ex \@minus -.2ex}%
                                       {2.3ex \@plus.2ex}%
                                       {\normalfont\fontfamily{phv}\fontsize{16}{19}\bfseries}}
    \renewcommand\subsection{\@startsection{subsection}{2}{\z@}%
                                         {-3.25ex\@plus -1ex \@minus -.2ex}%
                                         {1.5ex \@plus .2ex}%
                                         {\normalfont\fontfamily{phv}\fontsize{14}{17}\bfseries}}
    \renewcommand\subsubsection{\@startsection{subsubsection}{3}{\z@}%
                                        {-3.25ex\@plus -1ex \@minus -.2ex}%
                                         {1.5ex \@plus .2ex}%
                                         {\normalfont\normalsize\fontfamily{phv}\fontsize{14}{17}\selectfont}}
    \makeatother
	
	\usepackage{amsmath}
	\usepackage{graphicx}
	\usepackage{enumerate}
	\usepackage{xcolor}
	\usepackage[pagebackref=true,breaklinks=true,colorlinks,bookmarks=false]{hyperref}
	\usepackage{url} 
	\usepackage[section]{placeins}
        \usepackage{titling}

\begin{document}

        \renewcommand\maketitlehooka{%
        \setlength\parindent{0pt}%
        \begin{minipage}{\textwidth}
        \begin{center}
            Distribution Statement ``A'' (Approved for Public Release, Distribution Unlimited)
        \end{center}
        \end{minipage}\vskip 2.5ex
        \par
        }
        

		\def\spacingset#1{\renewcommand{\baselinestretch}%
			{#1}\small\normalsize} \spacingset{1}
		
		\if0\blind
		{
			\title{\bf Comprehensive Dataset of Face Manipulations for Development and Evaluation of Forensic Tools}
			\author{Brian DeCann (brian.decann@str.us), Kirill Trapeznikov (kirill.trapeznikov@str.us) \\
             STR, Woburn, Massachusetts, United States }
			\date{}
			\maketitle
		} \fi
		
		\if1\blind
		{

            \title{\bf \emph{IISE Transactions} \LaTeX \ Template}
			\author{Author information is purposely removed for double-blind review}
			
\bigskip
			\bigskip
			\bigskip
			\begin{center}
				{\LARGE\bf \emph{IISE Transactions} \LaTeX \ Template}
			\end{center}
			\medskip
		} \fi
		\bigskip
		
			

	\spacingset{1.5} 

\section{Introduction} \label{s:intro}

\begin{figure*}[t]
    \centering
    \includegraphics[width=5.95in]{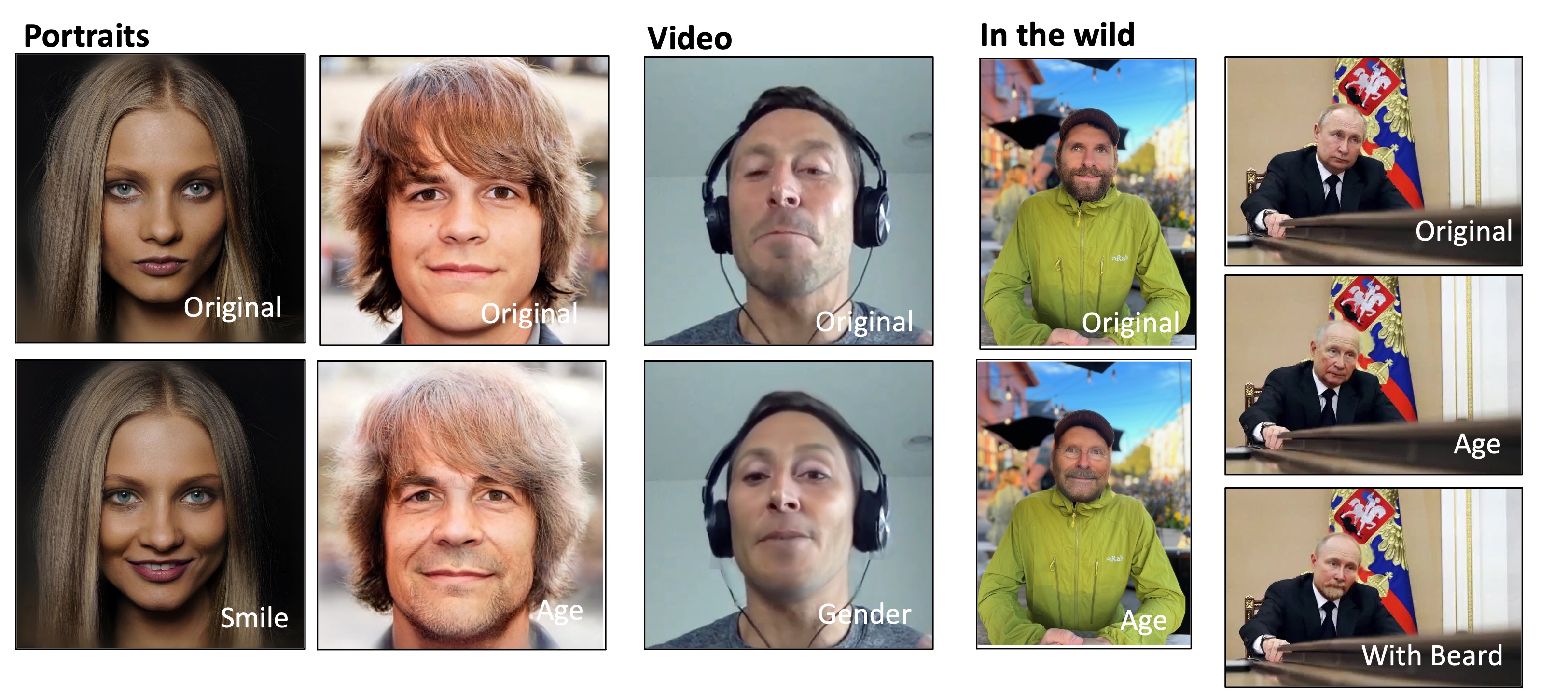}
    \caption{Examples of face manipulations in media. State-of-the-art models are capable of applying photorealistic manipulations in portrait-style (left), video (center), and in-the-wild media. Here, manipulations are generated using the approaches from Roich et al. \cite{roich2021pivotal}, and Tzaban et al. \cite{tzaban2022stitch}. }
    \label{fig:intro_example_manipulations}
\end{figure*}

Digital media (e.g., photographs, video) can be easily created, edited, and shared. Tools for editing digital media are capable of doing so while also maintaining a high degree of photo-realism. In other words, edits to digital media can be made to be unrecognizable to the human eye. Many edits to digital media are generally benign. For example, color-balancing or contrast-enhancement of photographs improves visual acuity and is aesthetically pleasing. However, edits can also be applied for malicious purposes. State-of-the-art face editing tools and software, for example, can artificially make a person appear to be smiling at an inopportune time, or depict authority figures as frail and tired in order to discredit individuals. At present, these editing models are generally based on StyleGAN \cite{roich2021pivotal}\cite{tzaban2022stitch}\cite{zhang21}\cite{yin2022styleheat}, although image diffusion approaches  \cite{dockhorn2021score}\cite{ramesh2022hierarchical}\cite{rombach2022high}\cite{xiao2021tackling}\cite{preechakul2022diffusion}\cite{dockhorn2021score} also perform very well. Additionally, NERF-based approaches \cite{gu2021stylenerf} have also been developed. These approaches are all generally well performing while being quite different from one another, illustrating a variety of methods a user could utilize. Examples of edited (facial) photographs are illustrated in Figure \ref{fig:intro_example_manipulations}. The example edits illustrated show a variety of semantic changes made to a face (e.g., neutral pose to smile and appearing older) in both controlled, portrait-style frontal face images as well as in-the-wild, full-scene images. 

Given the increasing ease of editing digital media and the potential risks from misuse, a substantial amount of effort has gone into media forensics. To this end, we created a challenge dataset of edited facial images to assist the research community in developing novel approaches to address and classify the authenticity of digital media. Our dataset includes edits applied to controlled, portrait-style frontal face images and full-scene in-the-wild images that may include multiple (i.e., more than one) face per image. The goals of our dataset is to address the following challenge questions: 

\begin{enumerate}
    \item Can we determine the authenticity of a given image (edit detection)?
    \item If an image has been edited, can we \textit{localize} the edit region?
    \item If an image has been edited, can we deduce (classify) what edit type was performed?
\end{enumerate} 

The majority of research in image forensics generally attempts to answer item (1), detection. To the best of our knowledge, there are no formal datasets specifically curated to evaluate items (2) and (3), localization and classification, respectively. Our hope is that our prepared evaluation protocol will assist researchers in improving the state-of-the-art in image forensics as they pertain to these challenges.

\FloatBarrier

\section{Face Manipulations in Portrait Images}

\begin{figure}[ht]
    \centering
    \includegraphics[width=5.0in]{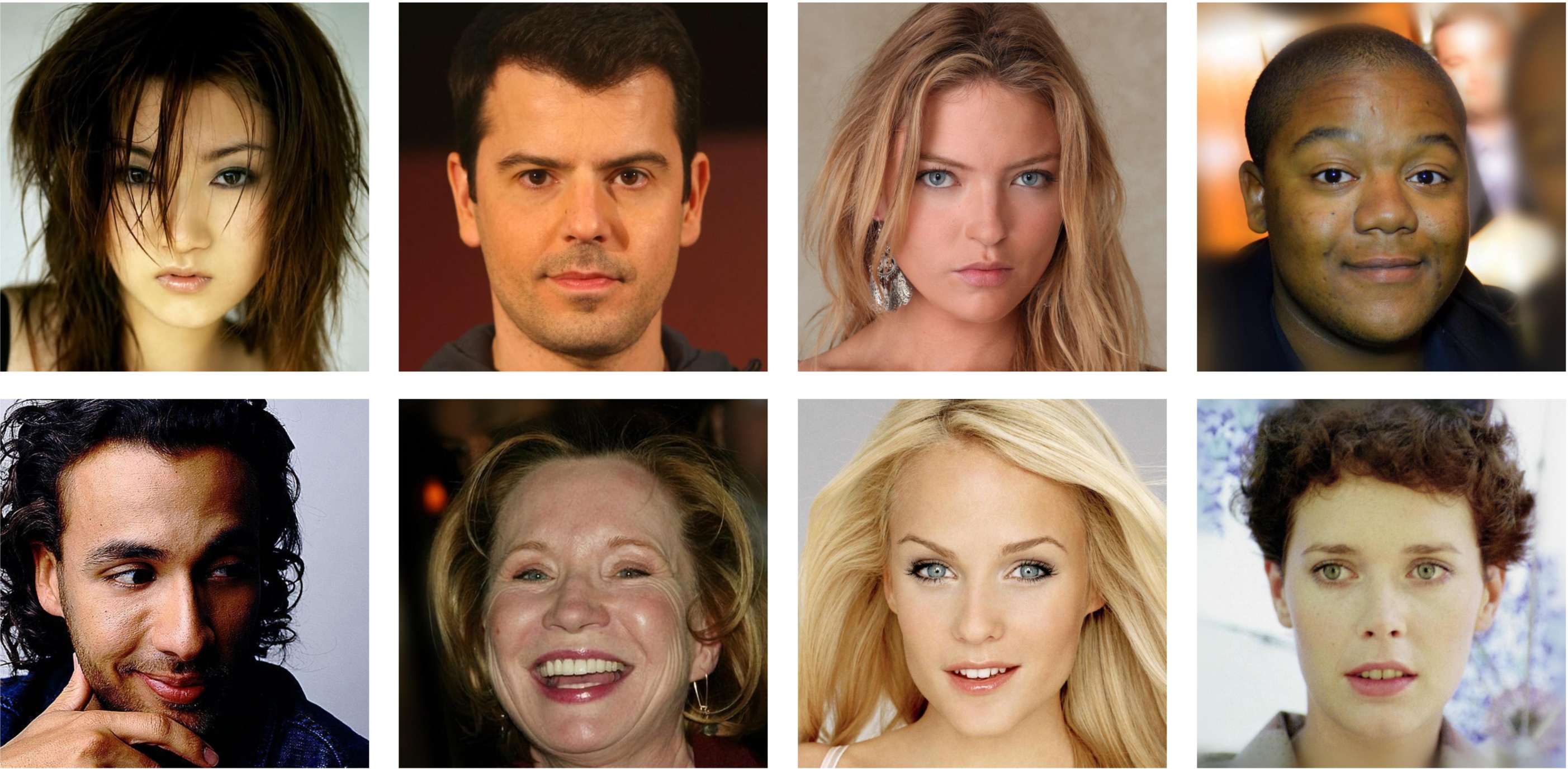}
    \caption{Examples of portrait-style images from the CelebA-HQ dataset \cite{CelebAMask-HQ}.}
    \label{fig:portrait_image_examples}
\end{figure}

A portrait image is a type of face image where a significant majority of the image foreground denotes a human face. Example portrait-style images are illustrated in Figure \ref{fig:portrait_image_examples}.

\subsection{Portrait Image Dataset}
\subsubsection{Dataset}
We compiled a dataset of edited portrait-style images. The image data was sourced from a subset of the CelebA-HQ dataset \cite{CelebAMask-HQ}. The CelebA-HQ dataset is a high-quality subset of the Large-Scale CelebFaces Attributes (CelebA) dataset \cite{liu2015faceattributes}. The CelebA-HQ dataset consists of 30,000 high-quality versions of images in the CelebA dataset. The images denote square-cropped faces from photographs captured in-the-wild and are saved at a resolution of 1024x1024 pixels (versions at 128x128 and 256x256 pixels also exist). In our subset, we only consider identities that appear at least twice (i.e., there are at least two images of a given identity) in the image data.

\begin{figure}[ht]
    \centering
    \includegraphics[width=5.97in]{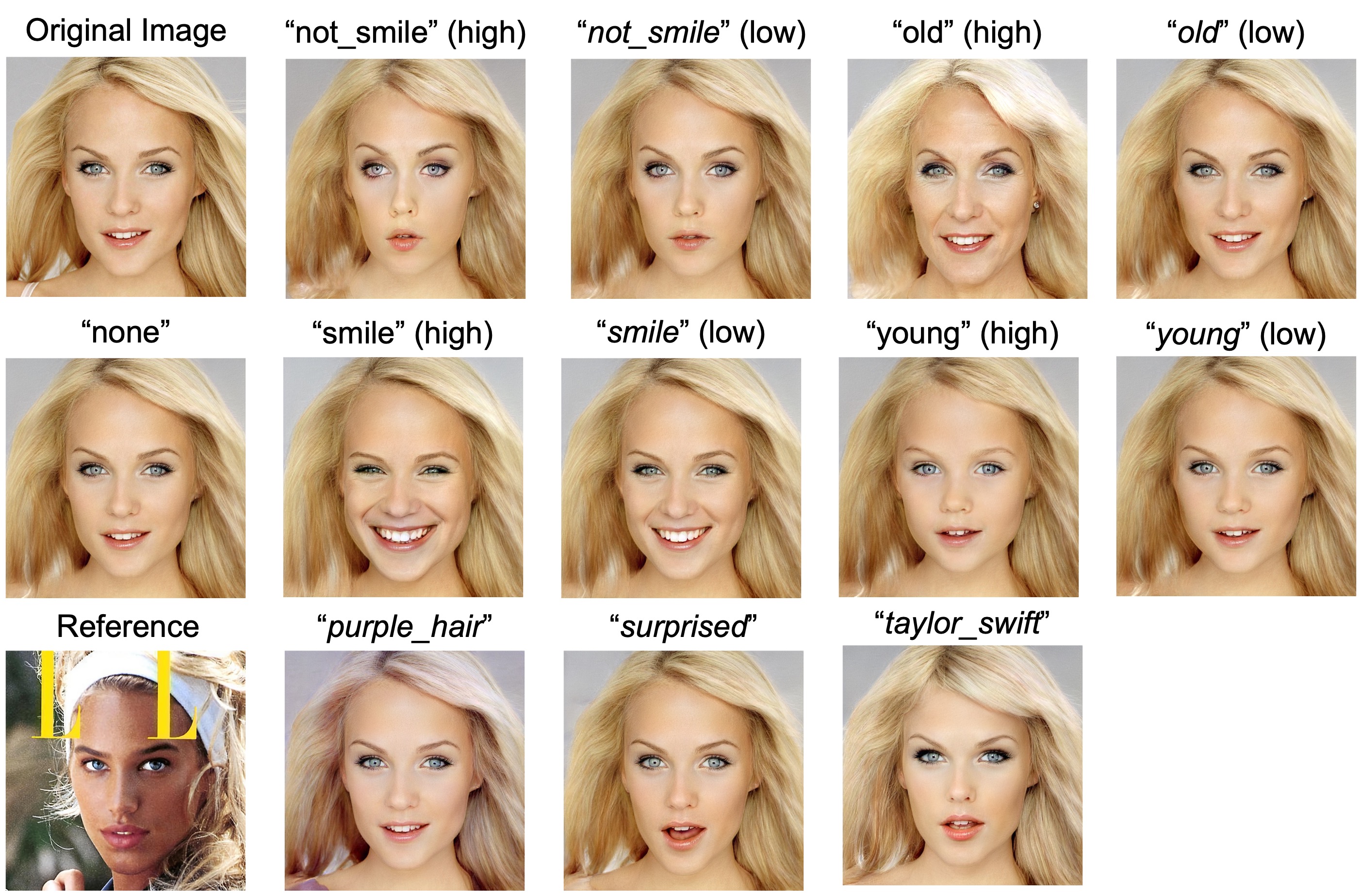}
    \caption{Examples of a set of images for a given subject in our manipulation dataset. Here, labels that are \textit{italicized} denote manipulations that exist strictly in the testing partition and are not present in the training partition.}
    \label{fig:portrait_image_set_examples}
\end{figure}

We created two partitions of image data for training and testing purposes. The training partition contains a total of 6,846 total images. Each sampled CelebA-HQ image in the training partition is manipulated in five (5) separate instances, in combination with the original (unedited) image. Each sampled CelebA-HQ image is also paired with a separate (unedited) image of the same face identity as a reference. The five manipulations consist of ``smile'' (smile added or enhanced), ``not smile'' (smile removed or reduced), ``young'' (face is modified to appear younger), ``old'' (face is modified to appear older), and ``surprised'' (face is modified to include a surprised expression). We applied the Pivotal Tuning approach by Roich et al. to create each manipulated image \cite{roich2021pivotal}. The testing partition contains a total of 7,644 images and includes the same types of manipulated images as in the training partition and an additional seven manipulated images for a total of twelve images per identity (plus the original and a reference). In the testing partition there are additional examples for ``smile'', ``not smile'', ``young'', and ``old'', where the edit magnitude is reduced. In addition, there are three novel manipulations not present in the training partition. These include ``purple{\_}hair'' (hair is modified to have a purple color), ``angry'' (face is modified to depict an angry expression), and ``Taylor Swift'' (face shape and features modified to appear similar to Taylor Swift). Each manipulation type is summarized in Table \ref{tab:portrait_image_set_edits}. An example of a subject with the set of applied manipulations is illustrated in Figure \ref{fig:portrait_image_set_examples}. In this example, each of the fourteen images would denote one subject in the testing partition. Labels that are \textit{italicized} would not appear for a given subject in the training partition. 

\begin{table}[ht]
    \centering
    \caption{Caption}
    \begin{tabular}{|c|l|}
        \hline
        \textbf{Edit Type} & \textbf{Remark} \\\hline
        ``smile'' & Smile added or enhanced \\\hline
        ``not smile'' & Smile removed or reduced \\\hline
        ``young'' & Face is modified to appear younger \\\hline
        ``old'' & Face is modified to appear older \\\hline
        ``surprised'' & Face is modified to depict a surprised expression \\\hline
        ``purple{\_}hair'' & Hair is modified to have a purple color \\\hline
        ``angry'' & Face is modified to depict an angry expression \\\hline
        ``Taylor{\_}Swift'' & Face shape and features modified to appear similar to Taylor Swift \\\hline
        
    \end{tabular}
    \label{tab:portrait_image_set_edits}
\end{table}

We remark that some identities may appear more than once in a given partition (training or testing), however an identity appearing in the training set will not appear in the testing set (and vice-versa). Both partitions are available in .png and .jpg format.

\begin{table}[ht]
    \centering
    \caption{Caption}
    \begin{tabular}{|c|c|c|}
        \hline
        & \textbf{Training Partition} & \textbf{Test Partition} \\\hline
        Image Count & 6846 & 7644 \\\hline
        Unique Edit Types & 5 & 8 \\\hline
        
    \end{tabular}
    \label{tab:portrait_image_set_counts}
\end{table}

\subsection{Evaluation Protocol}

For our portrait-style face manipulation dataset, we supply two challenges: detection and classification. A description of both challenges and associated outputs are described in the following sections.

\subsubsection{Detection}
The objective of the detection experiment is to identify whether a given image has been manipulated. For a given image in the testing partition return: 
\begin{itemize}
     \item (string) ``$<$filename$>$'' : Image filename
     \item (bool) [0,1] : Not edited or Edited
\end{itemize}

We measure balanced detection accuracy as the proportion of images that are correctly recognized as either edited or not edited.

\subsubsection{Classification}
The objective of the classification experiment is to classify the type of edit in a manipulated image. For a given image in the testing partition return: 
\begin{itemize}
     \item (string) ``$<$filename$>$'' : Image filename
     \item (string) ``pristine'' : \textbf{if} not edited; ``$<$edit{\_}type$>$'' : \textbf{ if} Edited
\end{itemize}

We measure classification accuracy as the proportion of edited images that are correctly recognized of being a given edit type.

\FloatBarrier

\section{Face Manipulation in the Wild}

An in-the-wild image can be described as an image where the principal foreground components (e.g., objects, people, animals) do not occupy the majority of the spatial image area and a large background is present. In-the-wild images are sometimes unconstrained in terms of lighting or camera angles. Unprocessed, or raw in-the-wild images can also vary greatly in terms of spatial resolution.

\subsection{In-the-Wild Image Dataset}

\subsubsection{Dataset}

\begin{figure}[ht]
    \centering
    \includegraphics[width=5.0in]{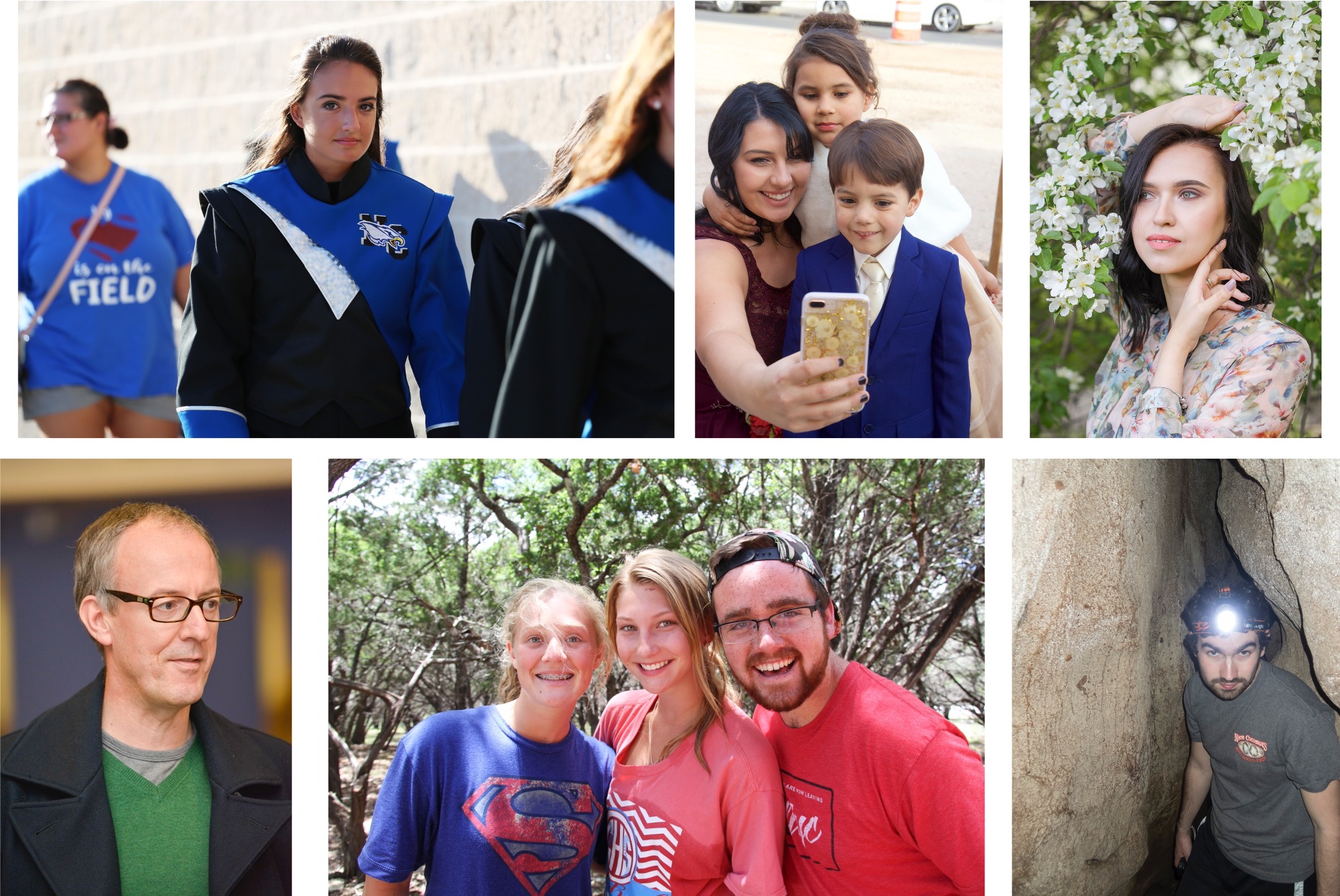}
    \caption{Examples of in-the-wild images from the FFHQ dataset \cite{karras2019style}.}
    \label{fig:inthewild_image_examples}
\end{figure}

We compiled a dataset of edited in-the-wild-style images. The image data was sourced from a subset of the Flickr-Faces-HQ (FFHQ) \cite{karras2019style}. The FFHQ dataset consists of 70,000 high-quality in-the-wild images. The authors of the FFHQ datset posit that the FFHQ data is much more variant in terms of age, ethnicity, background, and presence of facial covariates (e.g., eyeglasses, headwear) compared to CelebA-HQ. A version of the dataset consists futher of 70,000 detected, aligned, and cropped faces, which are saved at a resolution of 1024x1024 (a version at 128x128 also exists), but we only consider the raw, full-scene images. Example in-the-wild images from the FFHQ dataset are illustrated in Figure \ref{fig:inthewild_image_examples}.

Our edited in-the-wild dataset consists of a randomly sampled subset of the 70,000 raw in-the-wild FFHQ images. In our subset, we allow for the possibility that an image contains more than one person (face). This potentially adds an additional challenge in detecting and localizing edited faces. We created two partitions of image data for training and testing (validation) purposes. The training partition and test partition contain totals of 1,508 and 1,403 images, respectively. Within each partition, approximately 50\% of the images are edited, while the remaining images are ``pristine'' (i.e., not edited). In the training partition, 759 images are edited and 750 are pristine. For the testing partition, 652 images are edited and 750 are pristine. All images are saved in .jpg format with a randomly chosen quality factor in the set $Q_f \in [75,80,85,90]$. Summary information for the in-the-wild dataset is listed in Table \ref{tab:inthewild_image_set_summary}.

\begin{table}[ht]
    \centering
    \caption{In-the-wild face manipulation dataset summary}
    \begin{tabular}{|c|c|c|}
        \hline
        & \textbf{Training Partition} & \textbf{Test Partition} \\\hline
        Image Count & 1,508 & 1,403 \\\hline
        Pristine Images & 750 & 750 \\\hline
        Edited Images & 759 & 652 \\\hline
        Resolution & Variable & Variable \\\hline
        Image Format & .jpg & .jpg \\\hline
        
    \end{tabular}
    \label{tab:inthewild_image_set_summary}
\end{table}

\begin{figure}[ht]
    \centering
    \includegraphics[width=4.5in]{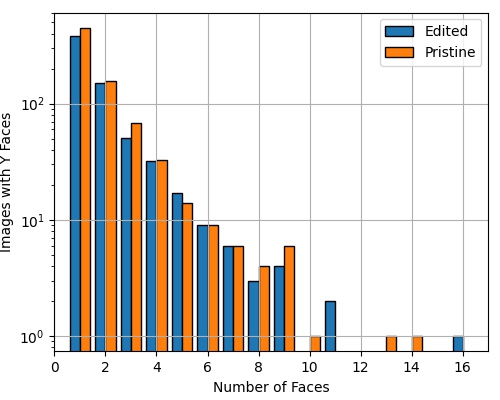}
    \caption{Face counts in our image set, separated by training and test partition. Note that the number of face counts is relatively well balanced for each partition and that a significant majority of images have one or two faces.}
    \label{fig:inthewild_face_counts}
\end{figure}

Unlike the portrait-style images, each edited image is only subject to a single edit type. In other words, there are not multiple copies of the same underlying image but with different edits applied. Images that are edited are subject to one of six possible manipulations. These include ``smile'', ``not{\_}smile'', ``young'', ``old'', ``male'', ``female''. We adopt the approach from Tzaban et al. to inject edits to in-the-wild images \cite{tzaban2022stitch}. 

\begin{table}[ht]
    \centering
    \caption{Types of image edits appearing in our in-the-wild face manipulation dataset}
    \begin{tabular}{|c|l|}
        \hline
        \textbf{Edit Type} & \textbf{Remark} \\\hline
        ``smile'' & Smile added or enhanced \\\hline
        ``not{\_}smile'' & Smile removed or reduced \\\hline
        ``young'' & Face is modified to appear younger \\\hline
        ``old'' & Face is modified to appear older \\\hline
        ``male'' & Face is modified to appear more masculine \\\hline
        ``female'' & Face is modified to appear more feminine \\\hline
    \end{tabular}
    \label{tab:inthewild_image_set_edits}
\end{table}

\begin{figure}
    \centering
    \includegraphics[width=5.97in]{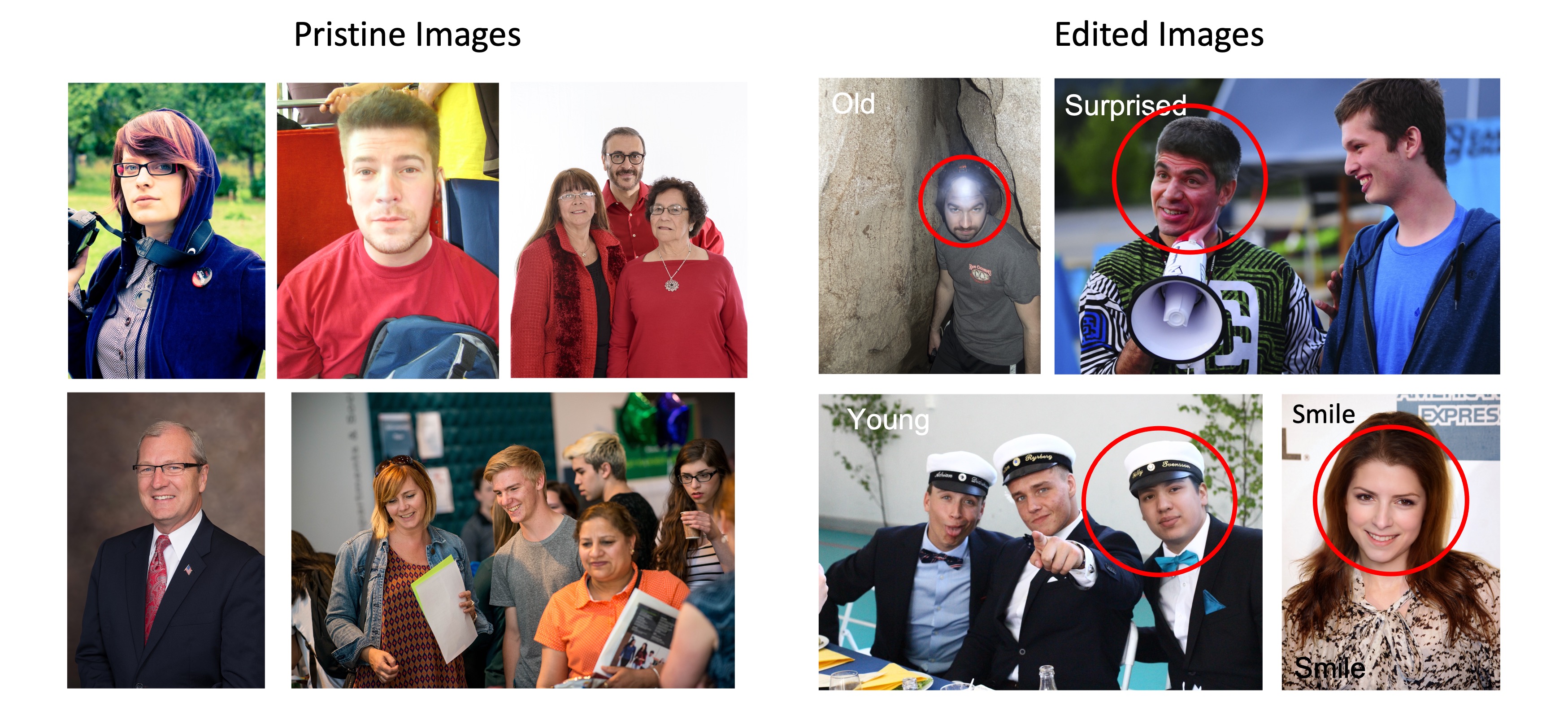}
    \caption{Examples of ``pristine'' (left) and ``edited'' (right) images in our in-the-wild face manipulation dataset.}
    \label{fig:inthewild_example_main}
\end{figure}

\begin{figure}
    \centering
    \includegraphics[width=5.97in]{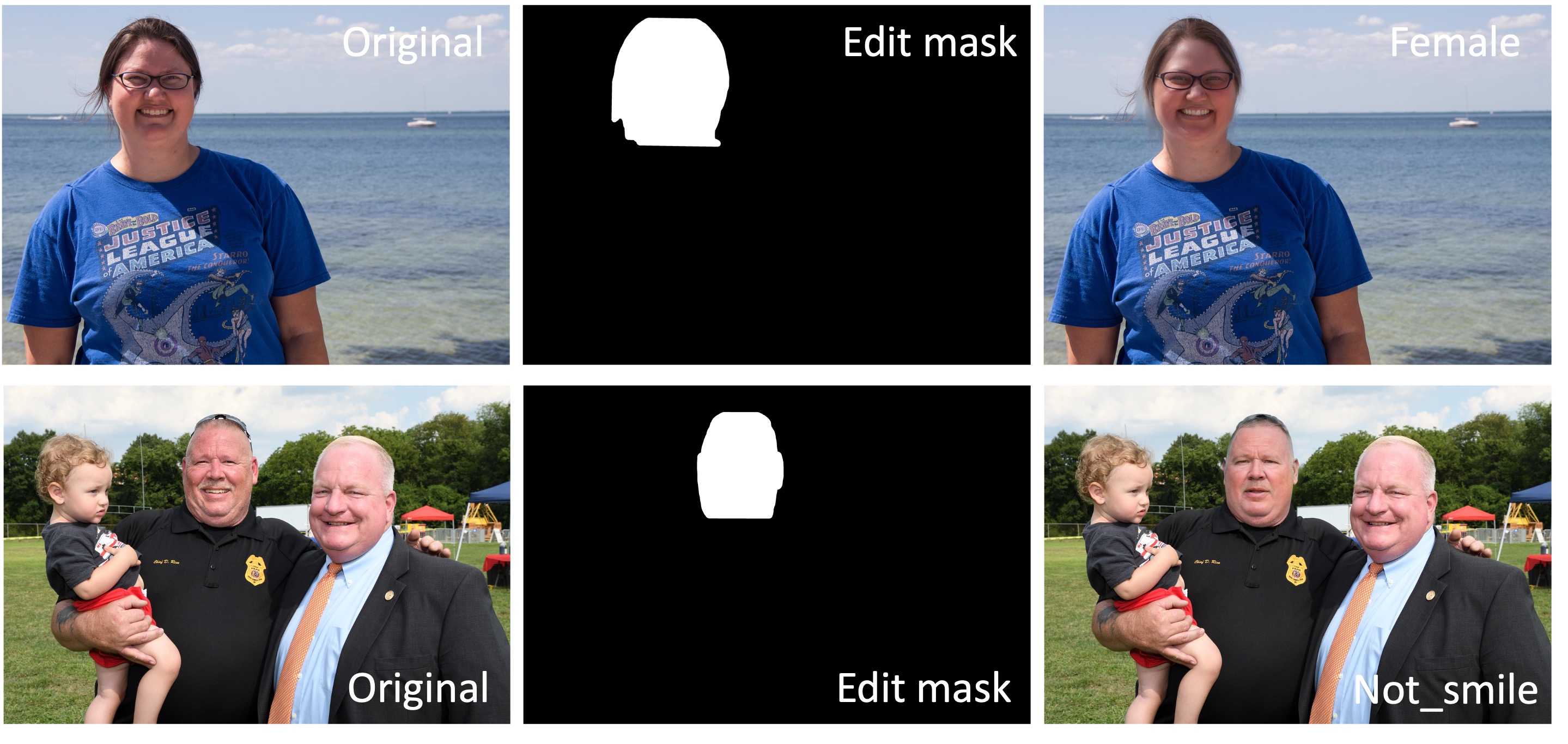}
    \caption{Examples of edit masks associated with edited in-the-wild images.}
    \label{fig:example_edit_masks}
\end{figure}

\begin{figure}[ht]
    \centering
    \includegraphics[height=3.0in]{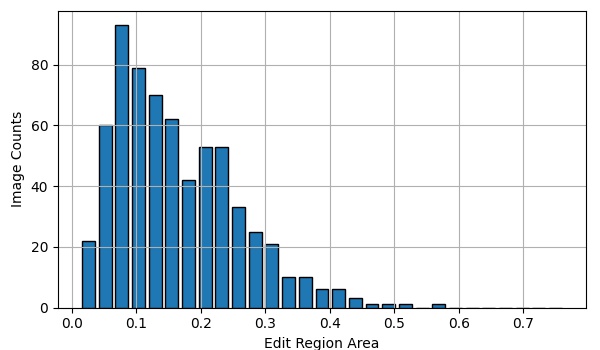}
    \caption{Proportion of edit region in our in-the-wild images. Note that the typical edit region size is approximately 10\% of the image.}
    \label{fig:inthewild_edit_area}
\end{figure}

For our in-the-wild face manipulation dataset, edits are localized to a region of the full-scene image. This is in contrast to the portrait-style face manipulation dataset, where images are fully synthesized from face-based GAN's. For the images in the in-the-wild face manipulation dataset that are edited, we also provide a binary mask that captures the spatial image region where the edit was performed and transplanted back into the image. The edit region is identified using a modified BiSeNet for faces \cite{yu2018bisenet}. Figure \ref{fig:example_edit_masks} illustrates examples of binary edit masks associated with edited in-the-wild images with one and multiple persons. Additionally, Figure \ref{fig:inthewild_edit_area} reports the proportion of the edit region for edited images. Typically, the size of the edit region is between 8\% and 20\% of the spatial image area.

\FloatBarrier
\subsection{Evaluation Protocol}
For our in-the-wild face manipulation dataset, we supply three challenges: detection, localization, and classification. A description of each challenge and outputs are described in the following sections.

\subsubsection{Detection}
The objective of the detection experiment is to identify whether a given image has been manipulated. For a given image in the testing partition return: 
\begin{itemize}
     \item (string) ``$<$filename$>$'' : Image filename
     \item (bool) [0,1] : Not edited or Edited
\end{itemize}

\subsubsection{Localization}
The objective of the localization experiment is to identify the specific image-region where an edit exists. For a given image in the testing partition, users must generate a binary mask, $\hat{M}$, denoting the estimated edit region. The estimated mask is compared against the ground truth, $M$ using Matthews Correlation Coefficient (MCC). The MCC (phi coefficient, or mean-square contingency coefficient) is a measure of association for binary variables. MCC is computed from the confusion matrix of the pixel-based binary estimations. This is mathematically described in Equation \eqref{eq:mcc}, where TP denotes True Positive ($M_k=\hat{M}_k=1$), TN denotes True Negative ($M_k=\hat{M}_k=0$), FP denotes False Positive ($M_k=0$, $\hat{M}_k=1$), and FN denotes False Negative ($M_k=1$, $\hat{M}_k=0$).

\begin{equation}
    MCC = \frac{TP\cdot TN - FP\cdot FN}{\sqrt{(TP+FP)(TP+FN)(TN+FP)(TN+FN)}}
    \label{eq:mcc}
\end{equation}

\subsubsection{Classification}
The objective of the classification experiment is to classify the type of edit in a manipulated image. For a given image in the testing partition return: 
\begin{itemize}
     \item (string) ``$<$filename$>$'' : Image filename
     \item (string) ``pristine'' : \textbf{if} not edited; ``$<$edit{\_}type$>$'' : \textbf{ if} Edited
\end{itemize}

In our in-the-wild face manipulation dataset the types of edits that are present in the training partition are also represented in the testing partition. Similarly, the types of edits that are in the testing partition are also represented in the training partition. Thus, the classification problem for this dataset is \textit{closed-set}. This is in contrast to the portrait-style data, where novel edit types exist in the testing partition. We encourage users utilizing this data and challenge problem to consider \textit{open-set} solutions as the set of potential edit types is near-unlimited.
\FloatBarrier

\medskip
\noindent {\bf Acknowledgement.}  This material is based upon work supported by the Defense Advanced Research Projects Agency (DARPA) under Contract No. HR001120C0129. The views, opinions and/or findings expressed are those of the author and should not be interpreted as representing the official views or policies of the Department of Defense or the U.S. Government. 

\bibliographystyle{ieee}
\spacingset{1}
\bibliography{ref}

\newpage
\appendix
\section{Additional Example Images}

In this section, we provide additional examples of manipulated face images from the portrait and in-the-wild image sets.

\subsection{Portrait Images}

\begin{figure}[ht]
    \centering
    \includegraphics[width=5.97in]{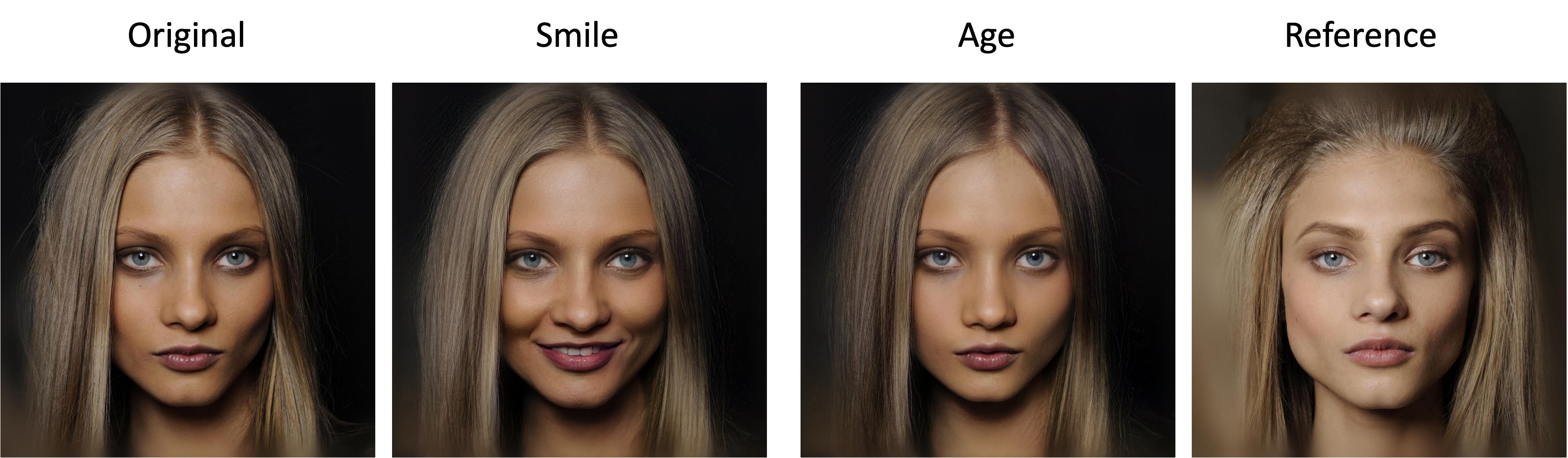}
    \caption{Manipulated portrait-style images with an added smile (middle left) and to appear younger (middle right). The right-most image denotes a reference (i.e., different image) for the same person.}
    \label{fig:example_portrait_00}
\end{figure}

\begin{figure}[ht]
    \centering
    \includegraphics[width=5.97in]{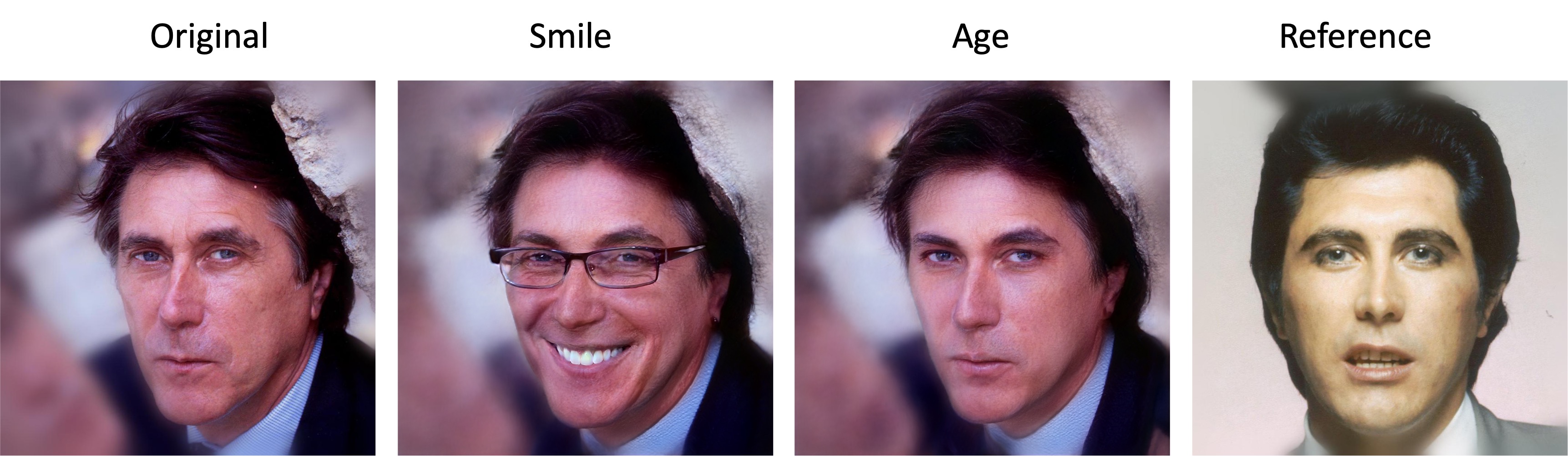}
    \caption{Manipulated portrait-style images with an added smile (middle left) and to appear younger (middle right). The right-most image denotes a reference (i.e., different image) for the same person.}
    \label{fig:example_portrait_01}
\end{figure}

\begin{figure}[ht]
    \centering
    \includegraphics[width=5.97in]{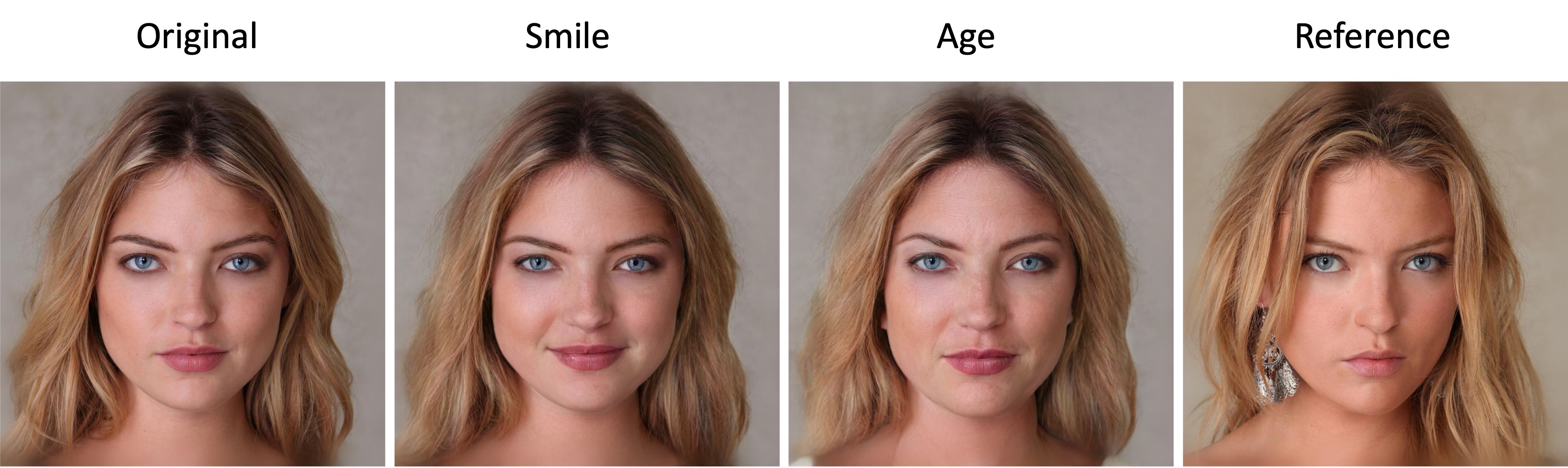}
    \caption{Manipulated portrait-style images with an added smile (middle left) and to appear younger (middle right). The right-most image denotes a reference (i.e., different image) for the same person.}
    \label{fig:example_portrait_02}
\end{figure}

Additional examples of edited portrait-style images are illustrated in Figures \ref{fig:example_portrait_00}-\ref{fig:example_portrait_02}.

\FloatBarrier
\subsection{In-the-Wild Images}
\begin{figure}[ht]
    \centering
    \includegraphics[width=5.97in]{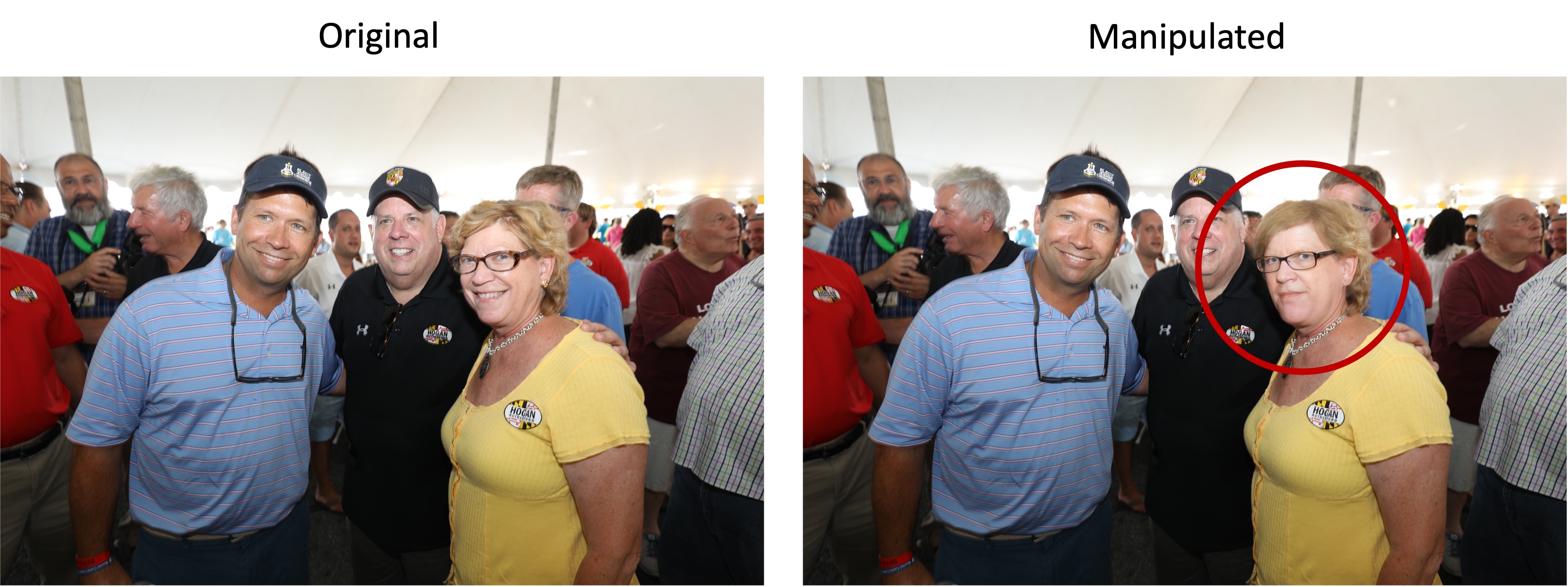}
    \caption{Manipulated in-the-wild image edited to remove smile.}
    \label{fig:example_itw_01}
\end{figure}

\begin{figure}[ht]
    \centering
    \includegraphics[width=5.97in]{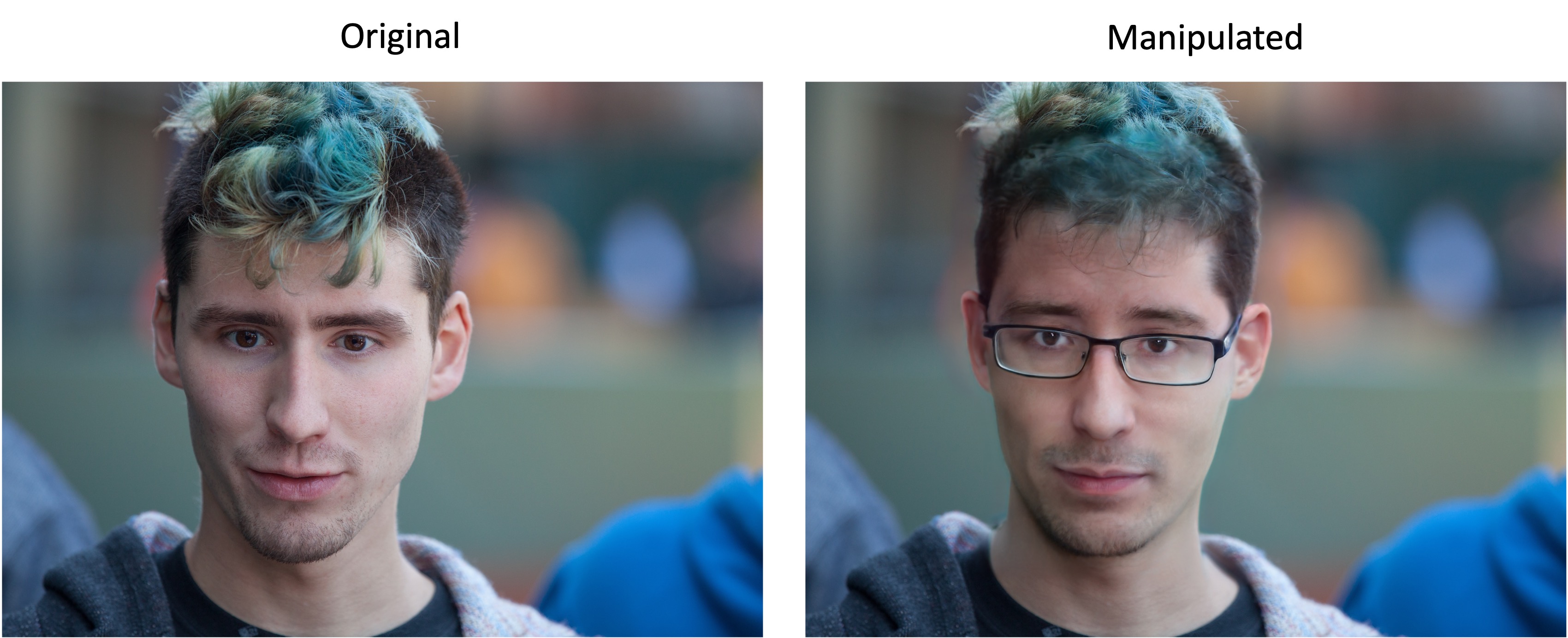}
    \caption{Manipulated in-the-wild image edited to add glasses}
    \label{fig:example_itw_02}
\end{figure}

Additional examples of edited in-the-wild images are illustrated in Figure \ref{fig:example_itw_01} and \ref{fig:example_itw_02}.

\end{document}